\documentclass[letterpaper, 10 pt, conference]{ieeeconf}  

\IEEEoverridecommandlockouts                              

\usepackage{graphics} 
\usepackage{epsfig} 
\usepackage{mathptmx} 
\usepackage{times} 
\usepackage{amsmath} 
\usepackage{amssymb}  
\usepackage{xcolor}
\usepackage{multirow}
\usepackage{adjustbox}
\usepackage{booktabs}
\usepackage{subcaption}
\usepackage{caption}
\usepackage{cite}
\usepackage{array}
\usepackage{siunitx}
\usepackage{symbols}
\usepackage{color,soul}
\usepackage{float}
\usepackage{stfloats}
\usepackage{graphicx}
\usepackage{hyperref}
\usepackage{tabularx}
\usepackage{subcaption}
\usepackage{newtxtext, newtxmath}

\title{\LARGE \bf
HyReach: Vision-Guided Hybrid Manipulator Reaching in Unseen Cluttered Environments
}

\author{Shivani Kamtikar$^{`*1}$, Kendall Koe$^{*1}$, Justin Wasserman$^{2}$, Samhita Marri$^{1}$, Benjamin Walt$^{1}$, \\
Naveen Kumar Uppalapati$^{1}$, Girish Krishnan$^{1}$, Girish Chowdhary$^{1}$
\thanks{*Equal Contribution, `Corresponding Author \newline
$^{1}$University of Illinois at Urbana-Champaign, $^{2}$Skild AI \newline
{\tt (skk7, kfkoe2, jbwasse2, marri2, walt, uppalap2, gkrishna, girishc)@illinois.edu}
}}

\begin{document}

\maketitle
\thispagestyle{empty}
\pagestyle{empty}

\begin{abstract}

\noindent As robotic systems increasingly operate in unstructured, cluttered, and previously unseen environments, there is a growing need for manipulators that combine compliance, adaptability, and precise control. This work presents a real-time hybrid rigid–soft continuum manipulator system designed for robust open-world object reaching in such challenging environments. The system integrates vision-based perception and 3D scene reconstruction with shape-aware motion planning to generate safe trajectories. A learning-based controller drives the hybrid arm to arbitrary target poses, leveraging the flexibility of the soft segment while maintaining the precision of the rigid segment. The system operates without environment-specific retraining, enabling direct generalization to new scenes. Extensive real-world experiments demonstrate consistent reaching performance with errors below \twocm across diverse cluttered setups, highlighting the potential of hybrid manipulators for adaptive and reliable operation in unstructured environments.

\end{abstract}

\section{Introduction}

Unstructured and cluttered environments present substantial challenges for robotic manipulation due to variability, occlusion, and constrained accessibility. In domains like agriculture, environmental monitoring, and disaster relief, robots must operate in complex scenes containing diverse obstacles, ranging from dense foliage to collapsed infrastructure. These settings often violate the assumptions of controlled environments~\cite{nazeer2023soft, nazeer2024rl}, demanding systems that adjust to clutter, avoid collisions, and perform goal-directed reaching with minimal prior knowledge of its surroundings. Many recent advances, however, target relatively uncluttered environments, such as tabletop settings~\cite{fang2024moka, wang2024push, huang2024toward}, where rigid manipulators with depth cameras struggle in tight or cluttered spaces due to limited dexterity and absence of passive compliance. Soft continuum arms (SCAs) provide greater adaptability and safe interaction~\cite{veil2025shape, kamtikar2022visual}, but their effectiveness is limited by a restricted workspace and challenges in accurate modeling, planning, and control.

Hybrid continuum manipulators offer a promising alternative, combining the reach and stability of rigid arms with the compliance and distal flexibility of SCAs~\cite{uppalapati2020valens, xu2024hybrid, koe2025learning}. These manipulators consist of rigid segments connected to a soft distal section. They have potential in scenarios requiring a unique combination of dexterity, adaptability, and precision, qualities that conventional rigid and soft manipulators struggle to achieve on their own. Recent studies have explored the design and modeling of hybrid manipulators~\cite{uppalapati2020valens, zhang2024design}, along with control and learning frameworks~\cite{koe2025learning, nazeer2024comparison, he2023modeling}. While some works address obstacle avoidance, they typically assume perfect prior knowledge of the environment or obstacles~\cite{xu2024hybrid, li2024s, veil2025shape}. Depth cameras can aid perception and scene modeling in novel environments; however, they are often impractical due to the limited payload capacity of SCAs. 

These limitations expose a critical gap in existing soft and hybrid manipulators: the lack of a unified, real-time framework that enables safe, goal-directed reaching in cluttered, previously unseen environments without relying on environment-specific modeling, calibration, or retraining. In particular, two key questions remain unresolved: (1) how hybrid systems can perceive and avoid obstacles while reasoning over their deformable geometry, and (2) how such capabilities can generalize across diverse unseen environments.

\begin{figure}[t]
\centering
\smallskip
  \includegraphics[width=0.44\textwidth]{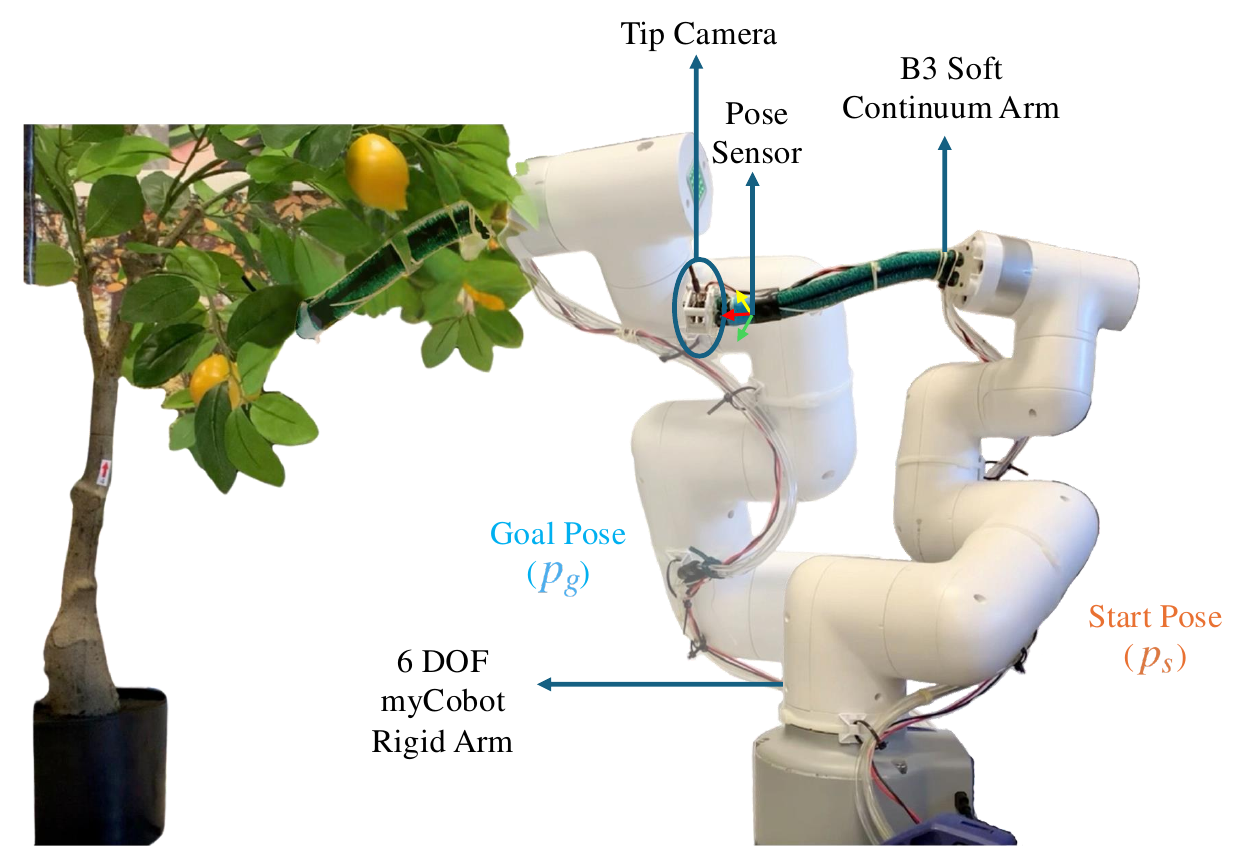}
 \caption{\textbf{Our system solves reaching tasks while using a hybrid rigid-soft continuum arm system.} Setup consists of a B3 (three bending actuators) soft continuum arm with a small RGB camera mounted on a 6DOF rigid manipulator. The setup also has a magnetic sensor (used only for data collection) that measures the pose of the end effector. We show two overlayed snapshots of the manipulator reaching toward a goal object through a cluttered environment.}
 \label{fig:system_fig}
 \vspace{-.6cm}
\end{figure}

To address this gap, we present a real-time, vision-guided hybrid manipulation framework that enables reliable object reaching in cluttered, unstructured, and previously unseen environments. The platform combines a standard 6-DoF industrial robotic arm with a tri-chambered bending (B3) SCA mounted at its distal end (Fig.~\ref{fig:system_fig}). The proposed framework couples multi-view reconstruction with shape-aware motion planning, explicitly reasoning over the deformable backbone of the hybrid manipulator to ensure safe and feasible motion, and pairs this with a learning-based controller for accurate execution in visually occluded, obstacle-rich scenes. We evaluate the framework across four increasingly challenging real-world environments, demonstrating reliable and robust performance under clutter and unseen scene geometry. This paper introduces the proposed framework, details its sensing, planning, and control components, and presents extensive real-world experiments that validate the feasibility of open-world, shape-informed hybrid reaching, while also discussing current limitations.

\noindent\textbf{Summary of contributions:}

\noindent\textit{1. Real-world demonstration of hybrid reaching.} To our knowledge, this is the first hybrid manipulator system capable of open-world object reaching in cluttered, unseen environments. The system achieves sub-2~cm accuracy and high success rates across multiple test environments without environment-specific fine-tuning.

\noindent\textit{2. Multi-view RGB reconstruction for obstacle-aware planning.} We develop a lightweight multi-view perception pipeline that enables obstacle-aware planning without relying on depth sensors or environment-specific retraining, making the system suitable for deployment on payload-limited hybrid manipulators.

\noindent\textit{3. Role of shape estimation in safe manipulation.} We demonstrate that explicitly incorporating shape estimation into the planning loop is critical for safe and reliable hybrid manipulation, significantly improving both safety and task success in cluttered environments.

\section{Related Work}

\begin{figure*}[htp!]
\centering
\smallskip
  \includegraphics[width=1\textwidth]{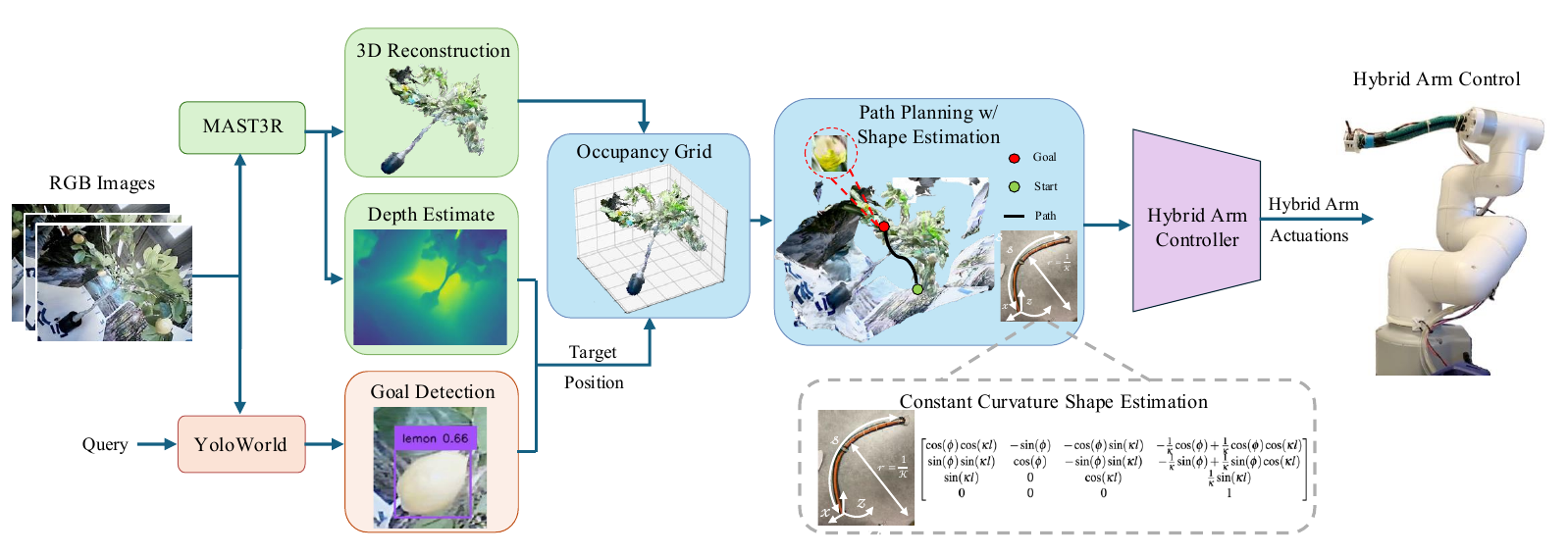}
  \caption{\textbf{Our pipeline for real-time reaching and control of a hybrid manipulator in complex, unstructured environments.} The pipeline comprises goal detection, 3D reconstruction, shape-informed path planning, and a learned controller for hybrid manipulators. 3D reconstruction, integrated with an occupancy grid, enhances scene understanding and identifies traversable areas. Shape-informed path planning optimizes paths by effectively navigating around obstacles. Additionally, our hybrid manipulator controller enables actuation to any arbitrary pose within the workspace.}
  \label{fig:pipeline}
  \vspace{-.5cm}
\end{figure*}

\textbf{Vision Guided Reaching.}
Goal-based reaching with rigid manipulators has been extensively studied in both structured and semi-structured environments as it supports core capabilities in robotic manipulation, from pick-and-place operations~\cite{hammoud2025online, wang2024genh2r} to assembly~\cite{yan2021high, xu2022noncontact} and inspection~\cite{ben2019automatic}. Recent vision-guided approaches leverage dense visual representations, such as multi-view reconstructions or point clouds, to enable reaching in cluttered scenes~\cite{zhu2024point, chisari2024learning}. These perception-driven techniques enable manipulators to reason about geometry and occlusion, improving safety and precision in motion planning.

Although existing learning-based frameworks have made significant progress in rigid manipulation, they typically rely on fixed kinematics, calibrated depth sensors, or known goal states~\cite{goyal2022ifor, chisari2024learning, christen2023learning}. In contrast, our system overcomes these limitations by leveraging low-cost RGB cameras for 3D reconstruction and geometry-aware planning, enabling real-time open-world reaching with hybrid, deformable kinematics - without requiring prior demonstrations or explicit goal poses.


\textbf{Hybrid and Soft-Arm Systems in Clutter.}
Planning in cluttered environments presents unique challenges for hybrid and soft manipulators, as their deformable geometries must navigate confined spaces while avoiding damage or excessive strain. Sampling-based algorithms such as RRT* and its variants~\cite{li2024s, meng2022rrt, luo2024efficient} have been applied to SCAs, but often assume full environmental knowledge or rely on third-person observations. More broadly, only a small number of studies have addressed hybrid manipulator planning~\cite{li2024s, luo2024efficient, huang2025grasping, koe2025learning}, and even fewer have reported consistent performance on real-world hardware~\cite{luo2024efficient, huang2025grasping}. Classical planners such as RRT* can effectively explore the configuration space and generate feasible paths. However, they depend on complementary modules such as shape sensing and modeling to handle the complex nonlinear dynamics of hybrid manipulators.

Frameworks such as SOFA~\cite{faure2012sofa}, Elastica~\cite{naughton2021elastica}, and Sorotoki~\cite{caasenbrood2024sorotoki} have enabled high-fidelity modeling of SCAs for tasks such as obstacle navigation, shape estimation, and shape control. However, these studies are typically confined to simulation and lack validation on real hardware, which limits their applicability in realistic, sensor-noisy environments. Shape estimation in real-world systems remains equally challenging. Although commercial fiber optic sensors provide accurate deformation feedback~\cite{galloway2019fiber}, their high cost restricts practical deployment. Recent research on SCA modeling and control has shown promise by leveraging learning-based frameworks~\cite{kamtikar2022visual, xu2024hybrid, nazeer2023soft, nazeer2024rl, koe2025learning}. However, these methods are typically validated in structured or obstacle-free settings, and their reliance on reinforcement learning or simulation-based training often leads to challenges in sim-to-real transfer and reward tuning.

In this work, we demonstrate a learning-based hybrid manipulator system that operates in cluttered, previously unseen real-world environments. By reasoning directly from a first-person visual perspective and accounting for the geometry of both rigid and soft segments, the system enables safe and adaptive motion generation in complex real-world settings.

\textbf{Hybrid-Arm Robot Control.}
Various numerical methodologies, including Piecewise Constant Curvature (PCC)~\cite{he2023modeling, xu2021visual}, Cosserat Rod Theory~\cite{shi2023position}, and Variable-Strain Model~\cite{9057619}, have been employed to model and control the dynamics of SCAs. For instance, He~\etal~\cite{he2023modeling} combined a PCC model with a data-driven module to control a hybrid manipulator. Similarly, Xu~\etal~\cite{xu2022visual} demonstrated a two-stage visual servoing strategy for pushing objects linearly with a hybrid system. Despite these advances, the inherently flexible, deformable, and high-DOF nature of soft and hybrid manipulators introduces significant complexity, making accurate modeling and control with traditional methods particularly challenging. 

Learning-based control offers a model-free alternative that has proven effective for purely soft systems~\cite{kamtikar2022visual, nazeer2023soft, nazeer2024rl}, as it eliminates the need for precise analytical modeling of hyper-redundant structures. Building on this paradigm, we employ a learning-based controller that enables our hybrid manipulator to reach arbitrary 6-DoF poses. This approach achieves precise control across a wide range of configurations without explicit system modeling, demonstrating the practicality of learning-based strategies for hybrid manipulation in real-world settings.


\section{Method}
To achieve reliable reaching with hybrid manipulators in cluttered environments, we propose a three-stage pipeline. First, the perception module integrates 3D scene reconstruction with open-world object detection, eliminating the common assumption that the target must be visible from the robot’s initial pose. Second, a shape-informed path planner leverages the 3D-reconstruction to compute obstacle-aware trajectories toward the goal pose. Finally, our learning-based controller executes these trajectories by actuating both rigid and soft segments of the manipulator to accurately reach the target.

\subsection{Perception}
The perception module's role within the pipeline is illustrated in~\figref{fig:pipeline}. It reconstructs the environment to provide a structured scene representation, forming the geometric basis for obstacle-aware planning and goal-directed control. The perception module begins by capturing RGB images from a monocular tip-mounted camera while the hybrid manipulator explores a small region around its home position to obtain multiple viewpoints. This multi-view strategy improves goal detection even when the target is initially occluded. The images are processed using Mast3r~\cite{leroy2024grounding} to estimate metric depth and reconstruct the scene. If the goal remains undetected, the manipulator performs additional exploratory motions to acquire new viewpoints, refining the reconstruction. This is challenging since depth cues are inferred solely from a low-resolution monocular stream, increasing depth-estimation uncertainty.

Goal objects are localized using YOLO-World~\cite{cheng2024yolo}, an open-world detector that enables natural-language queries and avoids dependence on fixed object categories. This design choice enables open-world flexibility and avoids dependence on a fixed set of object categories. YOLO-World is further chosen for its lightweight inference speed and ease of integration with real-time pipelines, making it well-suited for on-board robotic perception. Detected object locations are fused with the depth map to estimate 3D target positions. The target with the highest confidence is then paired with four candidate approach directions to generate corresponding candidate poses. 

The reconstructed point cloud is discretized into an occupancy grid encoding traversable and occupied regions. This grid serves as the environment model for the downstream shape-informed path planner, which generates safe trajectories for the hybrid manipulator to reach the target object. Details of the planning process follow in the next section.

\subsection{Path Planning with Shape Estimation}
Path planning is crucial for enabling hybrid manipulators to navigate around clutter and reliably reach a target, especially with occlusions. A key contribution of this paper is a shape-aware planning formulation that explicitly reasons over the deformable backbone of the soft segment during planning. Unlike conventional approaches that perform collision checking only at the end-effector~\cite{xu2024hybrid}, our planner evaluates collisions along the entire hybrid backbone and enforces asymmetric feasibility constraints: strictly collision-free motion for the rigid links while allowing bounded, controlled contact for the soft segment. This formulation exploits compliance as a planning primitive rather than treating deformation as a disturbance.

\noindent\textbf{Shape Estimation:} To incorporate the SCA's geometry into planning, we use a Constant Curvature (CC) model~\cite{hannan2003kinematics, rao2021model, wang2021survey}. Although CC-based modeling is less accurate than higher-fidelity alternatives (e.g., Cosserat rod solvers), it is computationally efficient and suitable for online planning. Estimating the SCA’s shape during path generation reduces the likelihood of collisions while maintaining real-time feasibility. Importantly, shape-informed planning ensures that candidate paths are not only kinematically valid but also safe with respect to the manipulator’s deformable geometry.

To compute the shape of the hybrid manipulator for path planning, the homogeneous transformation matrix of the shape of the soft distal link, $T_{s}$, is calculated as follows:

\begin{equation}
\resizebox{0.9\columnwidth}{!}{$
T_{s} =
\begin{bmatrix}
    R & t \\
    0 & 1
\end{bmatrix}
\begin{bmatrix}
    \cos (\phi)\cos (\kappa l) & -\sin(\phi) & -\cos(\phi)\sin (\kappa l) & 
    -\tfrac{1}{\kappa}\cos(\phi)(1- \cos(\kappa l)) \\
    \sin (\phi)\sin (\kappa l) & \cos(\phi) & -\sin(\phi)\sin (\kappa l) &
    -\tfrac{1}{\kappa}\sin(\phi)(1- \cos(\kappa l)) \\
    \sin(\kappa l) & 0 & \cos(\kappa l) &
    \tfrac{1}{\kappa}\sin(\kappa l) \\
    0 & 0 & 0 & 1
\end{bmatrix}
$}
\end{equation}

$R$ and $t$ represent the orientation and position of the rigid link to which the SCA is attached, $\kappa$ represents curvature of the SCA, $\phi$ represents rotation about the vector tangent to the SCA base, and $l$ is the arc length. The soft-arm shape is estimated by discretizing $l$ to compute poses along the manipulator. The rigid segment shape is determined using joint angles and known rigid body geometry. 

\noindent\textbf{Path Planning:} We implement a modified Rapidly-exploring Random Tree Star (RRT*) algorithm~\cite{karaman2011sampling} for path planning. The planner incrementally builds a tree by sampling random states, steering toward samples from the nearest node, and evaluating candidate edges for feasibility. The core modifications are:

\textit{1) Collision-aware expansion:} Candidate trajectories are evaluated against the occupancy grid by checking collisions along the entire arm backbone. To guarantee safe motion, the rigid segment must remain collision-free, while the soft segment is allowed limited contact within a predefined threshold.

\textit{2) Collision thresholding:} A tunable hyperparameter, $\tau$, defines the maximum allowable number of collisions along a candidate path. This threshold is empirically determined and allows tailoring the planner to different environments, for example, tolerating higher contact rates in deformable environments, and enforcing stricter constraints in environments with rigid or safety-critical obstacles. 
Let $q \in Q$ denote a candidate arm configuration, and let $P(q) = \{p_i\}_{i=1}^N , p_i\in \mathbb{R}^3 $ be a set of $N$ uniformly sampled points along the manipulator's backbone under the configuration $q$, obtained via shape estimation described above. $N$ was set to 50. Let the occupancy grid be represented by an indicator $occ:\mathbb{R}^3\rightarrow\{0, 1\}$, that returns 1 if and only if the point lies in an occupied voxel and 0 otherwise. The total collision count for configuration $q$ is defined as:
\begin{equation}
    C(q) = \sum_{i=1}^N occ(p_i)
\end{equation}
A configuration is considered feasible if and only if $C(q) \leq \tau$, where $\tau \in \{0,...,N\}$. The feasibility condition for the rigid component is given by $C_r(q) = 0$, and the feasibility condition for the soft segment is given by $C_s \leq \tau$, enforcing strict collision-free motion for the rigid section, while allowing limited allowable contact for the soft segment based on the predefined threshold. 

\textit{3) Dynamic shape updates:} The CC model is recomputed for each steering step to ensure that shape-dependent collisions are accurately represented during tree expansion.

Feasible paths are post-processed with a shortcutting heuristic~\cite{petit2021rrt} to remove redundant waypoints and produce smoother trajectories. If no solution is found within a maximum number of iterations, the planner terminates and returns failure, prompting replanning. The trajectory is generated before execution, and the waypoints are then passed to the controller. Combining occupancy-based feasibility with online shape estimation allows efficient and safe trajectory generation by leveraging the SCA’s compliance and limiting collisions. 

\begin{figure}[t]
\centering
\smallskip
 \includegraphics[width=0.45\textwidth]{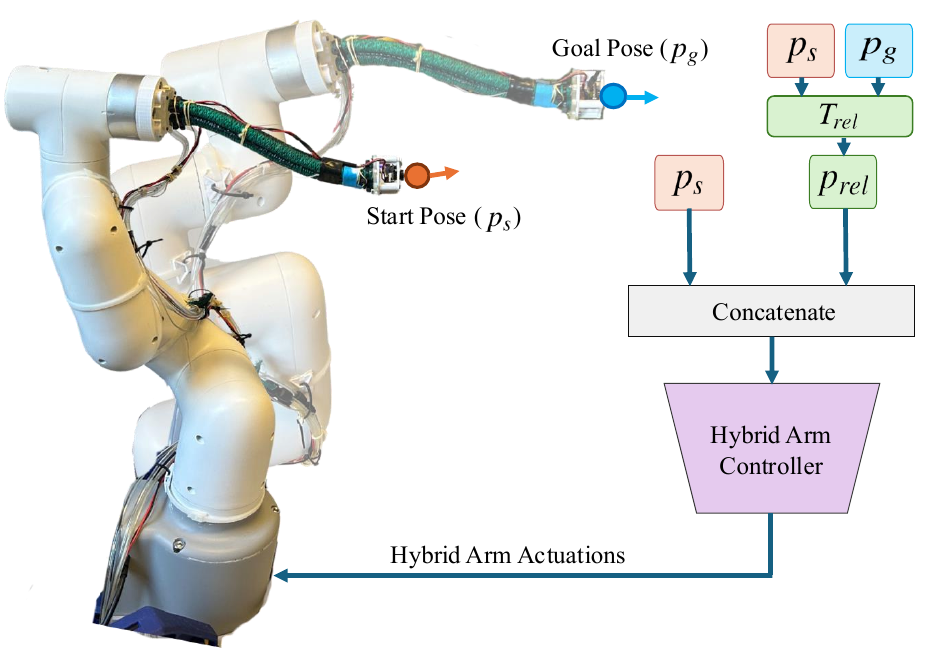}
 \caption{\textbf{The hybrid arm controller takes as input the start pose, $p_s$, and relative pose to the goal $p_{rel}$ and outputs hybrid arm actuations.} This fully learned controller successfully actuates the hybrid system to an arbitrary pose, avoiding the need for complex modeling of the hybrid system while enabling closed-loop control.}
 \label{fig:pose2act}
 \vspace{-.6cm}
\end{figure}

\begin{figure*}[htp!]
\centering
\smallskip
 \includegraphics[width=1\textwidth]{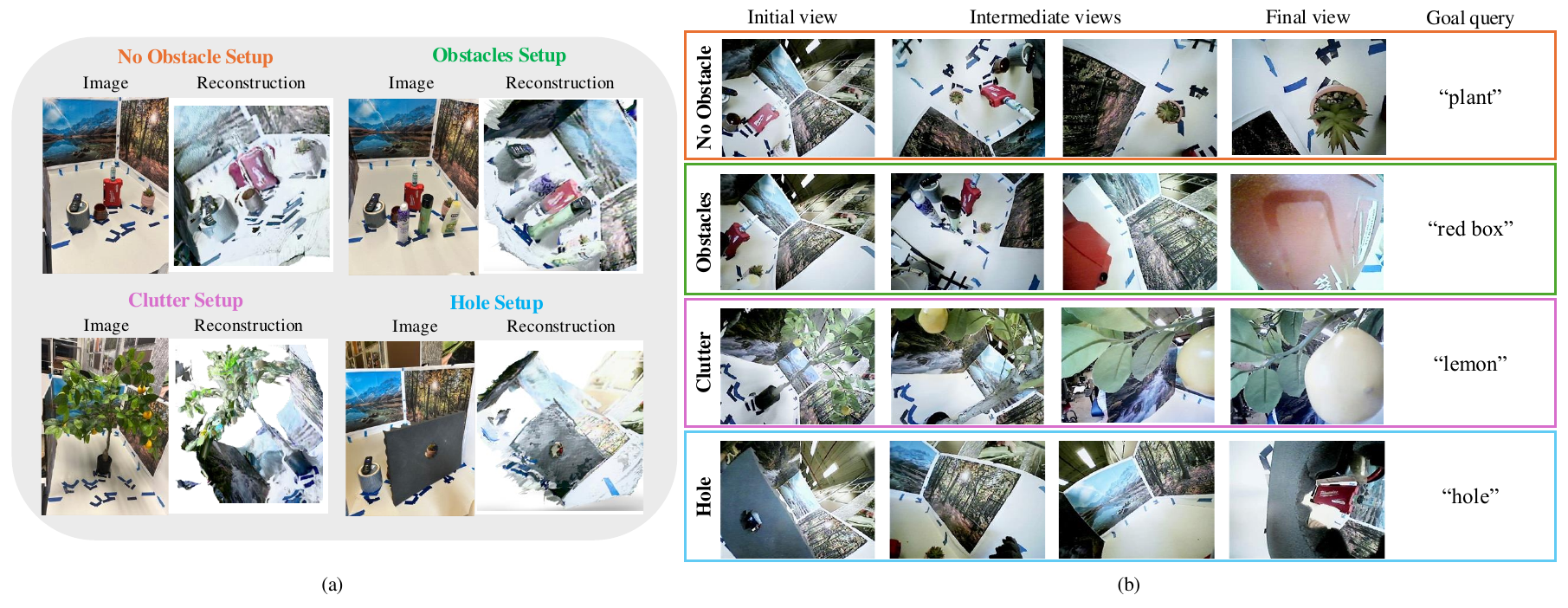}
 \caption{\textbf{(a) Experimental Setups:} The four experimental setups include \sceneopen, \sceneobstacle, \sceneclutter, and \sceneholes. An image of the setup, along with the reconstruction obtained from Mast3r~\cite{leroy2024grounding} is visualized. \textbf{(b) Experimental results (one example test run for each setup):} Shows initial view (start of the test), two intermediate views, and the final view (end of test) obtained from the tip camera. The final column shows the query/goal object that the hybrid arm was tasked to reach.}
 \label{fig:exp_results}
 \vspace{-.3cm}
\end{figure*}

\subsection{Learned Hybrid Arm Controller}
In previous work, it has been reported that controlling a hybrid manipulator is difficult due to its virtually infinite degrees of freedom, non-linear characteristics arising from material properties, coupling of rigid and soft segments, and the variety of designs and actuation techniques. To address this, we employ a learning-based closed-loop controller to actuate the hybrid manipulator (soft and rigid components) to a target pose. During deployment, the controller takes waypoints from the path planner and actuates the system incorporating the pose as feedback (\figref{fig:pose2act}). The controller inputs are the start pose and the relative pose to the goal, and the outputs are the soft-actuator commands and target joint angles for the rigid segment.

\noindent\textbf{Data Collection and Training.}
Data collection is performed by systematically incrementing each actuation dimension to ensure the hybrid manipulator explores its entire 9DOF workspace. For each configuration, the pose (position and orientation) and corresponding control inputs are recorded after oscillations settle, yielding $9536$ data points.

The model is trained using a mean-squared error (MSE) loss between the predicted actuations $\hat{u}$ and ground truth controls $u_{\textrm{gt}}$: 

\begin{equation}\label{Pose2Act Loss Function}
    L(x) = \frac{1}{||D||} \sum_{i = 1}^D ||\hat{u}^i - u^i_{\textrm{gt}}||^2
    \vspace{-.1cm}
\end{equation}
 
The input, $x$, is a concatenation of the current pose and the relative pose to the goal. The inputs were passed through a 20-layer MLP network designed with a bottleneck architecture and ResNet blocks to compute the desired normalized actuations. The model is trained with a batch size of 2000, learning rate of $1\times10^{-4}$, and hidden size of 15,000.



\begin{table*}[htp!]
\centering
\caption{\textbf{Across all environments, our method consistently outperforms the baselines in cluttered scenes.} 11 trials are performed for each setup. Intuitively, the success rate decreases with the complexity of the environment. The results show the importance of the perception pipeline and the shape-informed planner in successfully navigating obstacles. Furthermore, the data demonstrates the need for the soft segment in improving the success rate in cluttered, complex environments.}
\resizebox{0.99\textwidth}{!}{%
    \begin{tabular}{l@{\hskip 4mm} c c c c c c c c c}
    \toprule
    & & \multicolumn{2}{c}{\sceneopen} & \multicolumn{2}{c}{\sceneobstacle} & \multicolumn{2}{c}{\sceneclutter} & \multicolumn{2}{c}{\sceneholes}\\
    \cmidrule(lr){3-4}\cmidrule(lr){5-6}\cmidrule(lr){7-8}\cmidrule(lr){9-10}
    \# & Method & SR@2cm (\%) $\uparrow$ & SR@Touch (\%) $\uparrow$ & SR@2cm (\%) $\uparrow$ & SR@Touch (\%) $\uparrow$ & SR@2cm (\%) $\uparrow$ & SR@Touch (\%) $\uparrow$ & SR@2cm (\%) $\uparrow$ & SR@Touch (\%) $\uparrow$\\
    \midrule
    1 & Rigid only & \textbf{100.0} & \textbf{72.7} & 45.5 & 36.3 & 54.5 & 18.2 & 18.2 & 9.1 \\
    2 & Img2Act & 81.8 & 45.5 & 54.5 & 27.3 & 45.5 & 9.1 & 36.4 & 9.1 \\
    3 & \textbf{Ours} & 90.9 & 63.6 & \textbf{75.0} & \textbf{66.7} & \textbf{90.9} & \textbf{72.7} & \textbf{54.5} & \textbf{27.3}\\
    
    \bottomrule
    \end{tabular}}
    \label{tab:main}
\vspace{-.5cm}
\end{table*}

\section{Experiments and Results}
\subsection{Task Setup}
Experiments are performed in four test environments (Fig. \ref{fig:exp_results}(a)) with increasing difficulty that requires using the unique shape of the soft part of the manipulator. The physical setup description is given in the supplemental material. The first environment is \sceneopen, which acts as a basic table-top test environment. Next is the \sceneobstacle environment, where obstacles are placed in the environment to block the immediate path to the goal. Following this is the \sceneclutter environment, where a plant is placed in the scene. This requires the methods to be able to reach a target fruit on the plant even when faced with heavy occlusion. Finally,  the \sceneholes environment, where the goal is placed on the other side of a wall that contains a hole. The goal is fully occluded at the start of each episode, and the SCA needs to either look through the hole or around the wall to reach the goal. 

\noindent\textbf{Metrics:} Each trial is evaluated with three different success metrics. Here, SR refers to the success rate:
\begin{enumerate}
    \item \textit{SR@2cm}: The trial is a success if the end effector reaches within 2 cm of the query object and the object is in the line of sight. This metric is applicable for scenarios where the manipulator needs to observe and survey its surroundings. 
    \item \textit{SR@Touch}: The trial is a success if the end effector touches the object. This metric is used for scenarios where it is necessary to touch/interact with the target objects.
    \item \textit{Trans. Err}: Translation error with respect to the distance between the initial end effector position and the target position. This metric demonstrates precise reaching across larger distances and the range of motion of the system. 
\end{enumerate}

\noindent\textbf{Baselines:} As a baseline, we adopt a learned visual servoing approach~\cite{kamtikar2022visual} that maps images from a tip-mounted camera directly to actuator commands for the SCA. This method, \textit{Img2Act}, has robust performance in structured environments and serves as a representative visual servoing baseline. It was trained by collecting images in each experimental scene according to the data collection procedure of the prior work~\cite{kamtikar2022visual},  identical to that used for our controller. Separate models were trained for the four scenes using the same network architecture. The method requires environment access for training and a goal image at inference, limiting its ability to reach objects in unseen environments. The model was trained on 5184 image–actuation pairs for the \sceneopen setup, then fine-tuned with 1344 pairs for \sceneobstacle and another 1344 for \sceneholes. Pretrained weights enabled efficient adaptation through limited fine-tuning because \sceneobstacle and \sceneholes share similar visual features with \sceneopen. A separate model was trained for the visually distinct \sceneclutter setup using 5696 data pairs. Hyperparameters were tuned for each environment, yielding distinct, environment-specific models.


As another baseline, \textit{Rigid Only}, we evaluate a purely rigid manipulator by replacing the soft segment with a rigid link of equivalent length. This configuration preserves the overall kinematic reach but eliminates compliance, serving as a comparison that highlights the advantages of the soft segment in navigating cluttered environments. Baseline results are detailed in Table~\ref{tab:main} and~\ref{tab:trans.err}.

\noindent\textbf{Ablations:} To further evaluate our method, we perform the following ablation experiments.

\noindent\textbf{1) Effect of shape-informed planning:} We assess the contributions of the shape-informed planner within our pipeline. This is done by evaluating our pipeline with and without the shape-informed planner across all environments (11 trials for every environment), shown in Table~\ref{tab:shape}

\noindent\textbf{2) Effect of Collision Thresholds} We also study the effects of the collision threshold hyperparameter ($\tau$), comparing strict ($\tau = 0$), moderate ($\tau=4-6$), and lax ($\tau=10-15$) settings, which highlights the trade-off between safety (fewer collisions) and feasibility (higher path success rates). We compare results across all environments, with five trials per environment. Results are summarized in Table~\ref{tab:collisions}.

\noindent\textbf{3) Effect of controller input format on reaching accuracy:} We evaluate four input configurations to assess how state and goal representations affect controller performance: (1) current pose and transform to goal, (2) goal pose only, (3) current and goal poses, and (4) transform to goal only. Results are shown in Table~\ref{tab:pose2act}.

\subsection{Experimental Results}

\begin{table*}[htp!]
\centering
\caption{\textbf{Our method enables precisely reaching objects across larger distances than previous purely soft and hybrid manipulators.} Translation errors (Mean $\pm$ STD) measuring the distance from the end effector to the target object across all environments and methods are shown. Additionally, the initial distance from the target averaged across the 11 trials  (Init. $\Delta$ (cm)) is also given. }
\resizebox{\textwidth}{!}{%
    \begin{tabular}{l@{\hskip 4mm} c c c c c c c c c}
    \toprule
    & & \multicolumn{2}{c}{\sceneopen} & \multicolumn{2}{c}{\sceneobstacle} & \multicolumn{2}{c}{\sceneclutter} & \multicolumn{2}{c}{\sceneholes}\\
    \cmidrule(lr){3-4}\cmidrule(lr){5-6}\cmidrule(lr){7-8}\cmidrule(lr){9-10}
    \# & Method & Trans. Err (cm) $\downarrow$ & Init. $\Delta$ (cm) & Trans. Err (cm) $\downarrow$ & Init. $\Delta$ (cm) & Trans. Err (cm) $\downarrow$ & Init. $\Delta$ (cm) & Trans. Err (cm) $\downarrow$ & Init. $\Delta$ (cm)\\
    \midrule
    1 & Rigid Only & \textbf{1.0$\pm$1.2} & 33.0$\pm$2.8 & 7.6$\pm$6.2 & 31.7$\pm$2.0 & 5.2$\pm$5.5 & 31.2$\pm$1.9 & 9.8$\pm$8.1 & 31.5$\pm$1.8\\
    2 & Img2Act & 1.6$\pm$1.8 & 33.0$\pm$2.8 & 4.8$\pm$2.7 & 31.7$\pm$2.0 & 9.7$\pm$9.3 & 32.0$\pm$2.0 & 7.0$\pm$4.5 & 32.3$\pm$1.9\\
    3 & \textbf{Ours} & 1.2$\pm$1.7 & 33.2$\pm$2.0 & \textbf{1.6$\pm$2.3} & 32.5$\pm$2.0 & \textbf{1.0$\pm$1.9} & 31.3$\pm$1.9 & \textbf{2.9$\pm$2.4} & 31.6$\pm$1.8\\

    \bottomrule
    \end{tabular}}
\label{tab:trans.err}
\end{table*}

\begin{table*}[htp!]
\centering
\caption{\textbf{Shape estimation is essential for reliable operation in cluttered environments.} Our experiments demonstrate that incorporating shape-informed path planning is critical for generating feasible trajectories and avoiding severe collisions.}
\resizebox{0.99\textwidth}{!}{%
\begin{tabular}{l l *{12}{c}}
\toprule
& & \multicolumn{3}{c}{\sceneopen} & \multicolumn{3}{c}{\sceneobstacle} & \multicolumn{3}{c}{\sceneclutter} & \multicolumn{3}{c}{\sceneholes} \\
\cmidrule(lr){3-5}\cmidrule(lr){6-8}\cmidrule(lr){9-11}\cmidrule(lr){12-14}
\# & Method
& SR@2cm & SR@Touch & Trans. Err
& SR@2cm & SR@Touch & Trans. Err
& SR@2cm & SR@Touch & Trans. Err
& SR@2cm & SR@Touch & Trans. Err \\
& & (\%) $\uparrow$ & (\%) $\uparrow$ & (cm) $\downarrow$
& (\%) $\uparrow$ & (\%) $\uparrow$ & (cm) $\downarrow$
& (\%) $\uparrow$ & (\%) $\uparrow$ & (cm) $\downarrow$
& \%) $\uparrow$ & (\%) $\uparrow$ & (cm) $\downarrow$ \\

\midrule
1 & No-Shape
& \textbf{90.9} & \textbf{63.6} & 1.6$\pm$2.9
& 45.5 & 36.4 & 6.0$\pm$6.8
& 54.5 & 27.3 & 4.0$\pm$5.5
& 45.5 & 18.2 & 5.0$\pm$4.5 \\
2 & \textbf{Ours}
& \textbf{90.9} & \textbf{63.6} & \textbf{1.2$\pm$1.7}
& \textbf{75.0} & \textbf{66.7} & \textbf{1.6$\pm$2.3} 
& \textbf{90.9} & \textbf{72.7} & \textbf{1.0$\pm$1.9}
& \textbf{54.5} & \textbf{27.3} & \textbf{2.9$\pm$2.4} \\
\bottomrule
\end{tabular}
}
\vspace{-.3cm}
\label{tab:shape}
\end{table*}


\noindent\textbf{Result 1: Our hybrid manipulator system enables reliable object reaching in complex environments (Table~\ref{tab:main}) and~\ref{tab:trans.err}.}
Experiments demonstrate that our hybrid manipulator reliably performs open-world reaching in cluttered, unseen environments. These results highlight the manipulator's ability to reach target objects while safely navigating around obstacles. In the \sceneopen setup, it achieved 90.9\% for \textit{SR@2cm} and 63.6\% for \textit{SR@Touch}. As expected, success rates decline with increasing environmental complexity; however, our system maintains robust performance even in the most cluttered setting, achieving 90.9\% for \textit{SR@2cm} and 72.7\% for \textit{SR@Touch} in the \sceneclutter setup and 75\% for \textit{SR@2cm} and 66.7\% for \textit{SR@Touch} in the \sceneobstacle setup. The \sceneholes scenario, requiring precise motion through a narrow aperture, remains the most challenging: success rates reached 54.5\% (\textit{SR@2cm}) and 27.3\% (\textit{SR@Touch}), demonstrating feasibility in this highly constrained case. Representative successful trials across all environments are shown in Fig.~\ref{fig:exp_results}(b). 

For the \textit{Rigid Only} baseline, where the soft segment was replaced by a rigid link, the system performed well in \sceneopen (100\% and 72.7\%) but suffered frequent collisions in confined spaces, dropping to 45.5\% (\textit{SR@2cm}) in \sceneobstacle and 54.5\% in \sceneclutter. This is due to the absence of the compliance and dexterity provided by the soft continuum actuator, which are essential for navigating clutter and accessing partially occluded targets. Differences between \textit{SR@2cm} and \textit{SR@Touch} metrics arise from depth-estimation errors in Mast3R. Overall, results validate the use of a hybrid design to achieve robust reaching in unstructured, cluttered settings where existing methods fall short.

\noindent\textbf{Result 2: Multi-view reconstruction enables real-time, generalizable, and obstacle-aware reaching (Table~\ref{tab:main}).}   
Integrating multi-view reconstruction into the perception pipeline markedly improves performance and generalization. The \textit{Img2Act} baseline~\cite{kamtikar2022visual} performs well in simple, uncluttered scenes, achieving 81.8\% for \textit{SR@2cm} and 45.5\% for \textit{SR@Touch} in the \sceneopen setup, but fails to avoid obstacles and requires retraining for each environment. The lower performance on the \textit{SR@Touch} metric is primarily due to inaccuracies in the learned controller, which come from the non-linearities present in the SCA. In contrast, our system leverages 3D reconstruction for real-time reaching in unseen environments without retraining, explicitly reasoning about obstacles to ensure safe trajectories. These results confirm that combining perception, planning, and control yields reliable, generalizable reaching performance.

\begin{figure}[t]
\centering
\smallskip
 \includegraphics[width=0.35\textwidth]{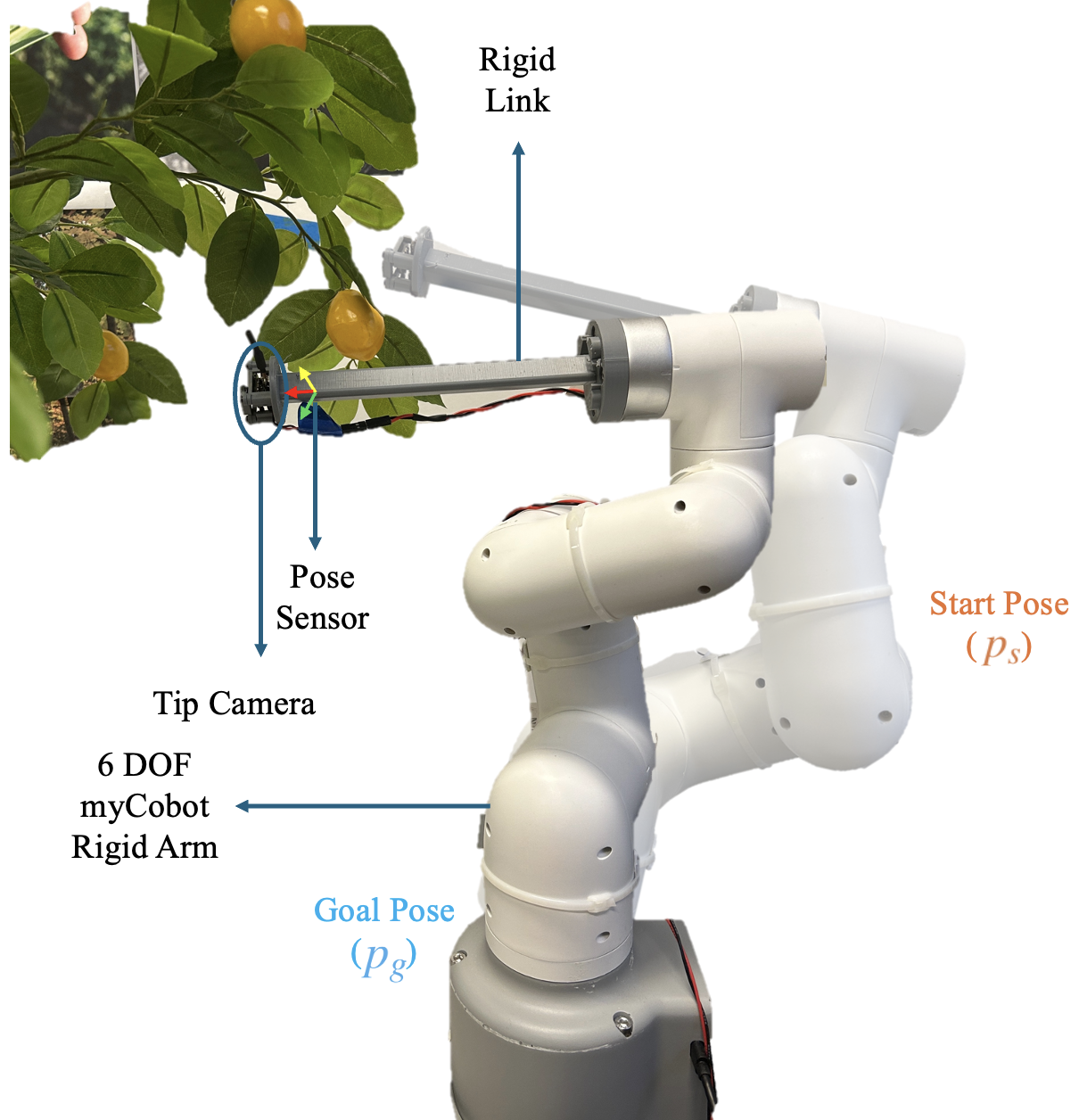}
 \caption{\textbf{\textit{Rigid Only} baseline hardware setup.} The soft segment in Fig. \ref{fig:system_fig} was replaced with an equivalent length rigid link.}
 \label{fig:rigid_link}
 \vspace{-.7cm}
\end{figure}

\noindent\textbf{Result 3: Shape estimation improves path safety and efficiency (Table~\ref{tab:shape})}. The \sceneopen setup is not significantly affected by SCA shape estimation with both methods achieving 90.9\% (\textit{SR@2cm}), as there were no collisions in this setup. However, in more complex scenarios, the absence of shape estimation significantly compromises both the safety of the environment and the SCA. In the \sceneobstacle setup, lack of shape estimation led to frequent, sometimes severe, collisions with obstacles, reducing the success rate to 45.5\% as compared to our method, which gave 75\%. While the arm often reached its target in the \sceneclutter setup, it occasionally became entangled with surrounding leaves and branches. Here too, our method achieved a success rate of 90.9\%, surpassing the 54.5\% success rate of the no-shape method. In \sceneholes, lack of shape estimation caused repeated collisions and overbending, reducing success to 45.5\%, while our method achieved 54.5\%. 

Although higher-fidelity shape estimation methods exist~\cite{zheng2024estimating, gruebele2021distributed}, our approximate model provides real-time performance with substantial gains in safety. Even with shape estimation, limited contact is permitted due to the inherent resilience of the SCA. Analysis of collision thresholds (Table~\ref{tab:collisions}) shows a clear trade-off between feasibility and safety: a strict threshold ($\tau=0$) prevents collisions but limits feasible paths, while a lax threshold ($\tau=10$–$15$) increases contact and reduces success. A moderate range ($\tau=4$–$6$) achieves the best balance, enabling safe interactions while maintaining high path feasibility in diverse environments.

\noindent\textbf{Result 4: Hybrid manipulators achieve precise reaching across extended workspaces (Table~\ref{tab:trans.err}).}
Our system enables accurate reaching over larger distances than prior soft-only~\cite{gan2022reinforcement} or hybrid systems~\cite{xu2024hybrid}. The mean translation error remains below 2~cm for \sceneopen, \sceneobstacle, and \sceneclutter setups, and below 3~cm for the challenging \sceneholes case. This expanded workspace makes hybrid manipulators more practical for applications such as agriculture and medical intervention, where soft systems are limited by reach.

\noindent\textbf{Result 5: The learned controller reliably actuates the hybrid manipulator to arbitrary 6-DoF goals.}
Given the current and relative goal poses, the controller outputs actuations for both rigid and soft components in a closed-loop fashion, achieving an average position error of 1.7~cm across all environments. Ablation experiments on the \sceneopen setup (15 trials for every input type), shown in Table~\ref{tab:pose2act}, confirm that using both current and relative poses yields the most accurate results among all input configurations tested.

\begin{table}[t]
    \centering
    \caption{\textbf{Effect of collision threshold on goal reaching} 
    Strict thresholds ($\tau=0$) yield safer but less feasible paths, while lax thresholds ($\tau = 10-15$) 
    come at the cost of increased collisions. Moderate threshold ($\tau=4-6$) strikes a good balance between safety and feasibility.}
    \setlength{\tabcolsep}{11pt}
    \begin{tabular}{lccc}
        \toprule
        Metric & $\tau=0$ & $\tau=10-15$ & $\tau=4-6$ \\
        \midrule
        SR@2cm & 47.7\% & 59.1\% & \textbf{77.8\%} \\
        SR@Touch & 29.5\% & 36.3\% & \textbf{57.5\%} \\
        Trans. Err (cm) & 5.9$\pm$5.6 & 4.1$\pm$4.8 & \textbf{1.6$\pm$2.0} \\
        \bottomrule
    \end{tabular}
    \label{tab:collisions}
    \vspace{-.2cm}
    
\end{table}

\begin{table}[t]
\centering
\caption{\textbf{Controller ablations} show that giving the current pose and the relative pose between the current and target pose improves prediction of the target actuations}
\setlength{\tabcolsep}{14pt} 
\begin{tabular}{c c c}
\toprule
Input Type & Trans. Err (cm) $\downarrow$ \\
\midrule
\textbf{Current Pose + Relative Pose to Goal} & \textbf{1.2$\pm$1.7} \\
Goal Pose & 9.1$\pm$7.8 \\
Current Pose + Goal Pose & 11.4$\pm$7.3 \\
Relative Pose to Goal & 15.9$\pm$10.5 \\
\bottomrule
\label{tab:pose2act}
\end{tabular}
\vspace{-.5cm}
\end{table}


\section{Limitations and Future Work}
The experiments demonstrate that open-world reaching of unseen objects in cluttered environments is achievable with a hybrid manipulator. However, the controller does not include an explicit objective to minimize control effort or actuation changes. Consequently, the predicted actuation commands reflect the learned data distribution rather than an energy-optimal solution. Even though this doesn't impact the achieved reaching accuracy, incorporating effort-aware regularization could improve trajectory efficiency. Another important direction for future work is evaluating payload-dependent behavior, enabling hybrid manipulators to support load-bearing tasks through load-aware planning. Dynamic environments with moving obstacles or targets are also not considered, which would require continuous replanning. Finally, future work could explore how obstacles might be leveraged, not just avoided, for reaching tasks.

\section{Conclusion}
We present a vision-guided framework for hybrid rigid–soft manipulators operating in cluttered, unstructured environments. The system integrates multi-view 3D reconstruction and shape-informed path planning for reliable obstacle avoidance, paired with a learning-based controller that accurately actuates the manipulator to arbitrary poses. Experiments show consistent improvements over baselines in both obstacle avoidance and target reaching, particularly in settings where rigid manipulators fail. These results highlight the effectiveness of combining shape-aware, perception-driven planning with learning-based control for robust real-world deployment of hybrid continuum manipulators.





\bibliographystyle{IEEEtran}{}
\bibliography{references}

\end{document}